\documentclass{article}

\usepackage[final]{airov24-arxiv}

\usepackage[utf8]{inputenc} 
\usepackage[T1]{fontenc}    
\usepackage{hyperref}       
\usepackage{url}            
\usepackage{booktabs}       
\usepackage{nicefrac}       
\usepackage{microtype}      
\usepackage{xcolor}         
\usepackage{graphicx}
\usepackage{tabularx}
\usepackage{multirow} 
\usepackage{amsfonts,amsmath,amssymb,amsthm}
\usepackage{tikz}
\newcolumntype{C}{>{\centering\arraybackslash}X}
\newcolumntype{R}{>{\raggedleft\arraybackslash}X}

\usepackage{algorithm2e}
\usepackage{algpseudocode} 
\RestyleAlgo{ruled}
\renewcommand{\Re}{\operatorname{Re}}

\title{Understanding the Convergence in\\Balanced Resonate-and-Fire Neurons}

\def\uzl{
Adaptive AI Lab\\
Institute of Robotics \& Cognitive Systems\\
University of Lübeck, Germany
}

\def\cwi{
Machine Learning Group\\
Centrum Wiskunde \& Informatica (CWI)\\
Amsterdam, The Netherlands
}

\author{%
Saya Higuchi\\
\uzl{}\\
\small\texttt{saya.higuchi@student.uni-luebeck.de}
\And
\And 
Sander M. Bohté\\
\cwi{}\\
\small\texttt{sbohte@cwi.nl}
\And Sebastian Otte\\
\uzl{}\\
\small\texttt{sebastian.otte@uni-luebeck.de}
}

\begin{document}

\maketitle

\begin{abstract}
Resonate-and-Fire (RF) neurons are an interesting complementary model for integrator neurons in spiking neural networks (SNNs). Due to their resonating membrane dynamics they can extract frequency patterns within the time domain. While established RF variants suffer from intrinsic shortcomings, the recently proposed balanced resonate-and-fire (BRF) neuron marked a significant methodological advance in terms of task performance, spiking and parameter efficiency, as well as, general stability and robustness, demonstrated for recurrent SNNs in various sequence learning tasks. One of the most intriguing result, however, was an immense improvement in training convergence speed and smoothness, overcoming the typical convergence dilemma in backprop-based SNN training. This paper aims at providing further intuitions about how and why these convergence advantages emerge. We show that BRF neurons, in contrast to well-established ALIF neurons, span a very clean and smooth---almost convex---error landscape. Furthermore, empirical results reveal that the convergence benefits are predominantly coupled with a divergence boundary-aware optimization, a major component of the BRF formulation that addresses the numerical stability of the time-discrete resonator approximation. These results are supported by a formal investigation of the membrane dynamics indicating that the gradient is transferred back through time without loss of magnitude.
\end{abstract} 

\section{Introduction}
Spiking neural networks (SNNs), with their biological plausibility and potential energy efficiency compared to conventional artificial neural networks (ANNs), have gained significant interest in recent years \citep{pfeiffer2018deep}. Furthermore, recent methodological advancements in recurrent SNNs (RSNNs) modeling rich and complex synaptic relations with external recurrencies, show potential for effective time-series learning \citep{bellec2018long, yin2021accurate,fang2021}. The spiking neuron model most prominently used for large-scale SNNs is the adaptive leaky integrate-and-fire (ALIF) neuron that models spiking of neurons based on the synaptic signals that integrate into the membrane potential with current leakage and an adaptive threshold \citep{bellec2018long}. Despite their simple and energy efficient nature, SNNs with ALIF neurons suffer from the convergence dilemma as well as limited memory capacity. The convergence dilemma highlights the slow and unstable convergence of the SNNs to learn essential representations of the data. It stems from the use of backpropagation through time (BPTT) algorithm \citep{werbos1990backpropagation} or its variant that updates the parameters based on the error signal propagated over the whole sequence, which potentially leads to the exploding or vanishing gradient problem \citep{kag2021training}. Combined with the limited memory capacity to associate recent and past events, it remains a challenge to control the flow of information and for the SNNs to learn effectively. Recent advancements in spiking neuron models have addressed the aspect of memory capacity by implementing ALIF neurons with trainable time-constants \citep{yin2021accurate,fang2021}, and gated LIFs \citep{yao2022glif}, as well as with a two-compartment model (LSTM-LIF; \citeauthor{zhang2023long}, \citeyear{zhang2023long}) for short- and long-term memories. Nonetheless, the convergence dilemma persists for the adaptive neurons, requiring many epochs for the RSNNs to converge properly \citep{yin2021accurate,fang2021,zhang2023long}.  

An interesting spiking neuron model is the resonate-and-fire (RF) neuron \citep{izhikevich2001resonate}, which is one of the most simple and efficient resonator models, but was less noticed in practical terms so far. It shows oscillatory membrane dynamics that extract frequencies in the time-domain and propagates this information to interconnected neurons through periodic spikes, enabling more efficient representation of information than conventional integrator models \citep{izhikevich2001resonate}. It can furthermore control information flow, implicitly indicating a flexible and potentially high memory capacity. However, intrinsic limitations, such as fluctuating gradients, excessive spiking, and absence of a tailored reset mechanism have hindered effective learning in RF networks. The recently proposed balanced RF (BRF) neuron alleviates these limitations and exhibits high task performance for time-series learning with considerable spike and parameter efficiency \citep{higuchi2024}. 
More notably, it displays remarkably stable and fast convergence during optimization, overcoming the convergence dilemma. We aim to gain an intuition and further understand the reasons underlying the emergence of such convergence advantage. 

In this study, the comparison of the RF neuron with and without the divergence boundary showed that the former is essential for the convergence of the BRF network. The smooth and near-convex error landscape of the BRF network suggests higher generalization capacity compared to vanilla RF or conventional ALIF neurons, empirically clarifying the reason underlying the stable convergence. Furthermore, the combination of the near-identity membrane state matrix and the divergence boundary that ensures spectral radius unity or below were found to be the key contributors for this phenomenon.

\section{Balanced resonate-and-fire neurons}

The BRF neuron \citep{higuchi2024} was proposed in two flavors: the Izhikevich variant following the formulation in \citet{izhikevich2001resonate} and the harmonic variant based on \citet{alkhamissi2021deep}. While both variants are somewhat comparable, the Izhikevich variant behaved consistently more stable and was in most of the cases more spike-efficient. Hence, we only focus on this variant here.

\begin{figure}[h]
    \centering
    \begin{tikzpicture}[]
    \node[anchor=south west] (image) at (0,0) {\includegraphics[width=0.5\linewidth]
    {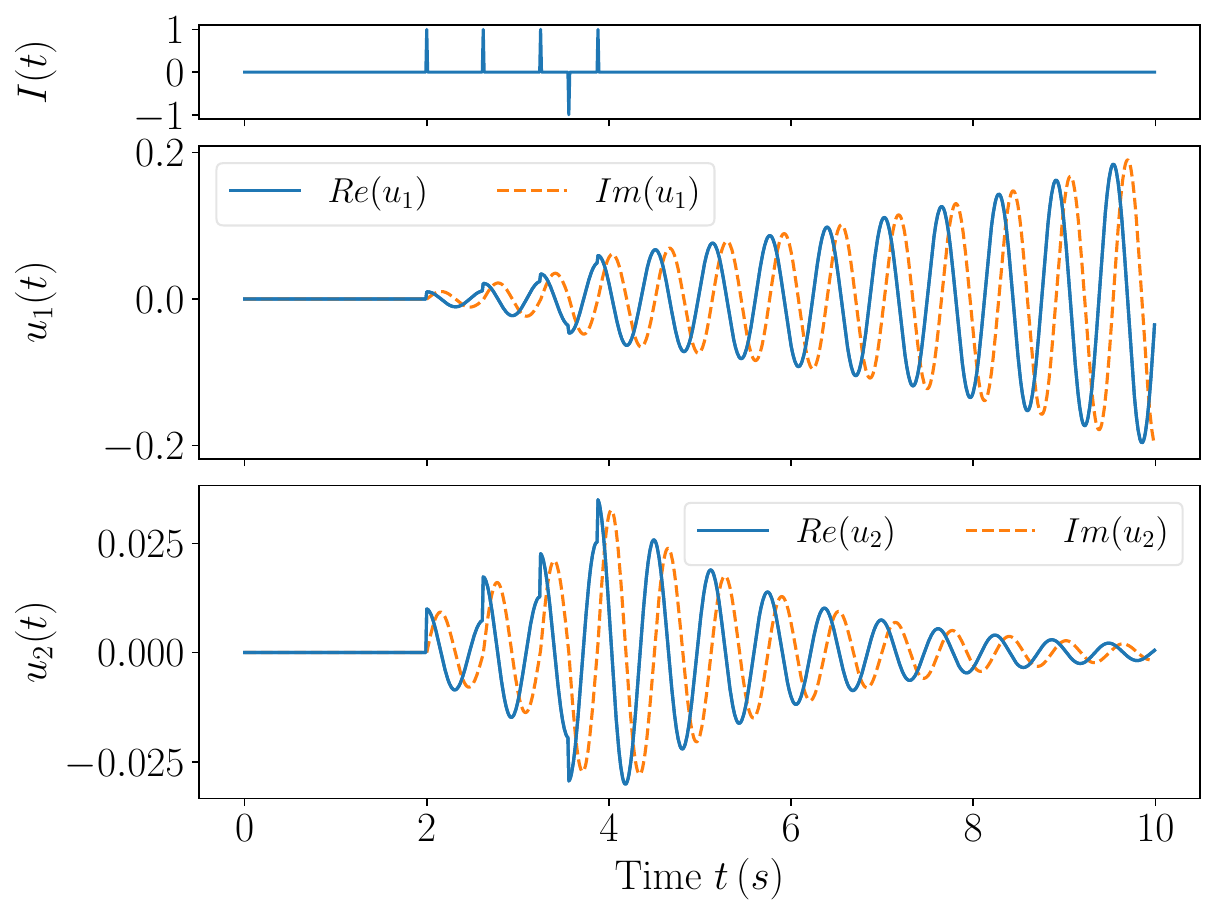}};
    \node[font=\scriptsize\linespread{0.9}\selectfont,align=left] at (2.47, 3.2) {unstable \\ ($\omega{=}10$, $b{=}-0.3$)};
    \node[font=\scriptsize\linespread{0.9}\selectfont,align=left] at (2.35, 1.21) {stable \\ ($\omega{=}10$, $b{=}-1$)};
  \end{tikzpicture}
    \includegraphics[width=.475\linewidth]{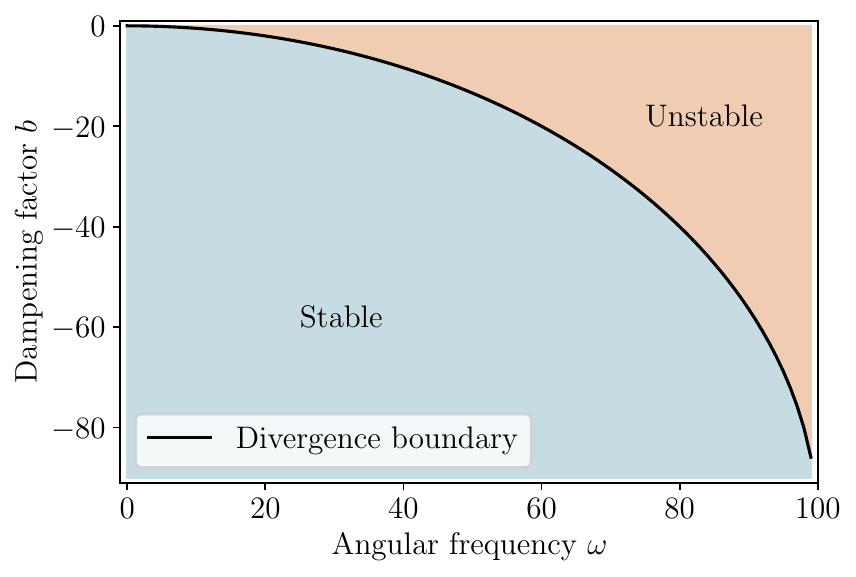} 
    \caption{Left: RF neuron behavior when current injected. Divergence (above) and convergence (below) shown with angular frequency of 10 and dampening of -0.3 and -1 \citep{higuchi2024}. Right: Parameter space of a single BRF neuron with $\delta = 0.01$. Combinations of $\omega$ and $b$ below the divergence boundary leads to convergence.}
    \label{figure:simulations}
\end{figure}

The oscillatory behavior of the membrane potential in an RF neuron is formulated via complex linear first-order differential equations: 
\begin{equation}\label{equation:2.1}
    \dot{u} = (b + i \, \omega) \, u  + I
\end{equation}
with $u \in \mathbb{C}$ and $I$ the injected current \citep{izhikevich2001resonate}, later replaced by $x$ to denote net input to the neuron. 
The angular frequency, represented by $\omega > 0$, tells us how fast a neuron oscillates in terms of radians per second. The dampening factor, denoted by $b < 0$, controls how quickly the neuron returns to the resting state. The behavior of the RF neuron is shown in \autoref{figure:simulations} (left).  

The Euler integration on \autoref{equation:2.1} with a time scale of $\delta$ gives:
\begin{equation}\label{eq:mem_update}
     u(t) = u(t-\delta) + \delta \, ((b + i\, \omega) \, u(t-\delta ) + x(t))
\end{equation}

This time-discrete update equation approximates the membrane dynamics of the RF neuron. The neuron's output spike $z(t)$ is now simply calculated by:
\begin{equation}
z(t) = \Theta \, (\Re(u(t)) - \vartheta_c)
\end{equation}
where $\vartheta_c$ defines the threshold of the neuron and $\Theta$ denote the Heaviside function. In the original formulation, the membrane potential is reset by setting the real part to the resting state and the imaginary part to 1 \citep{izhikevich2001resonate}.
This, however, did not work well in a practical learning setup. 
The BRF neuron extends the vanilla RF model in three aspects: (i) it adds a refractory period, (ii) it adds a smooth reset, a reset mechanism tailored for the characteristics of RF neurons, (iii) it includes a divergence boundary that helps to remain within parameter regime ensuring stable and converging resonator dynamics.

\subsection{Refractory period} 

The refractory period $q(t)$ is coupled with the spiking behavior and it temporarily increases the threshold to hinder frequent spiking: 
\begin{align}
\vartheta(t) &= \vartheta_c + q(t)\\
z(t) &= \Theta \, (\Re(u(t)) - \vartheta(t))\\
q(t) &= \gamma q(t-\delta ) + z(t - \delta )
\end{align}
%

where the refractory period decays exponentially with time. The default refractory period constant is $\gamma = 0.9$. 

\subsection{Smooth reset}
The smooth reset alleviates another limitation of the basic RF model, the traditional reset mechanism, that reduces the amplitude but disrupts the oscillation.
The smooth reset temporarily increases the dampening of the amplitude to decay faster after the neuron fires \citep{higuchi2024}:
\begin{equation}
b(t) = b_c - q(t)
\end{equation}
with $b_c$ the constant dampening factor. 
The smooth reset leads to a soft but sufficient decrease in the amplitude without disrupting the oscillation with sharp and abrupt decays.

\subsection{Divergence boundary}

A further drawback that arises from the numerical approximation of the continuous resonator dynamics, is the sensitivity to certain $\omega,~b_c$ combinations that leads to divergence behavior of the membrane potential, observed in \autoref{figure:simulations} (left). Due to its ever-increasing amplitude even without incoming signals, divergent RF neurons exhibit continuous spiking irrelevant to the task, increasing the load of the model to filter relevant information.

\begin{figure*}[b!]
    \centering
    \includegraphics[width=0.99\linewidth]{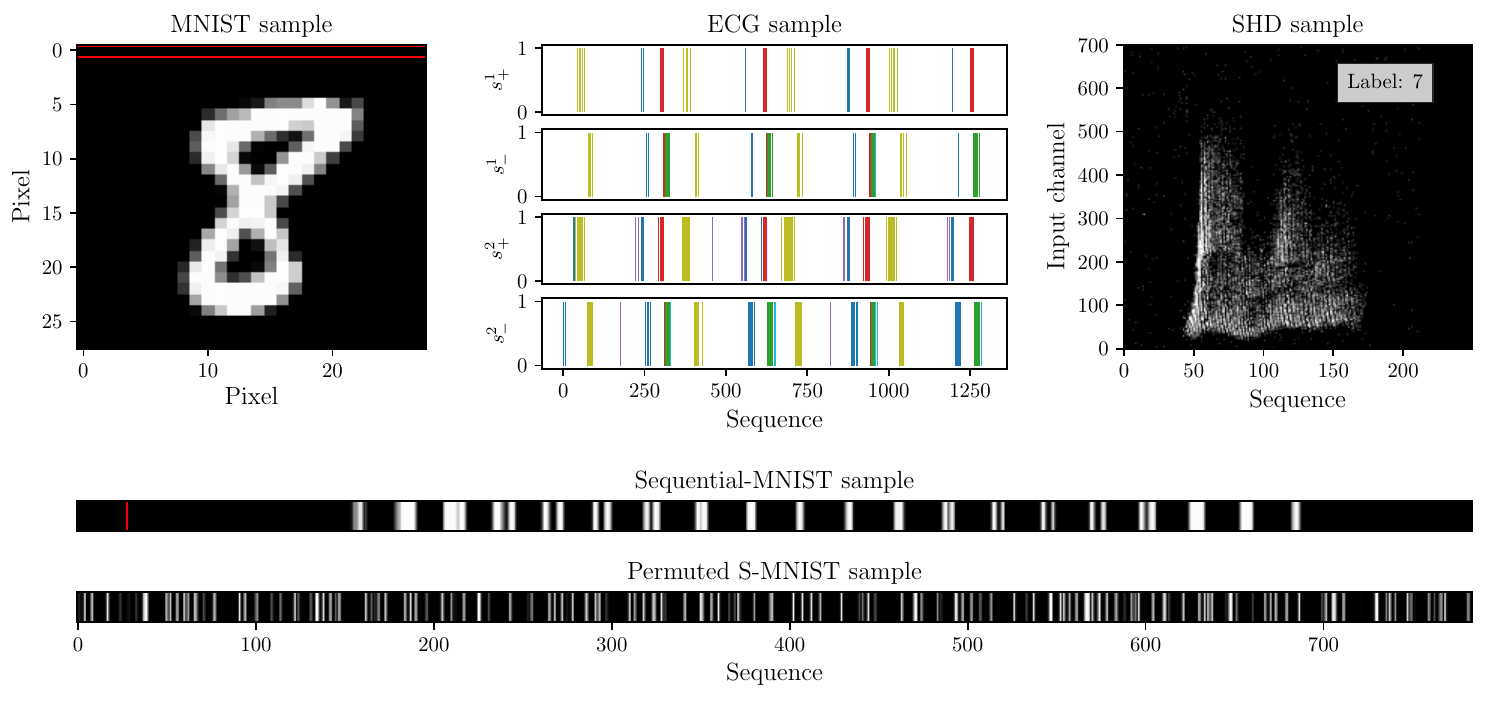}
    \caption{Overview over the datasets. Examplary MNIST image and its corresponding sequential and permuted representations. Common pixel row outlined in red on MNIST and S-MNIST sample. ECG sample after level cross encoding. SHD sample after preprocessing. Adapted from \cite{higuchi2024}.}
    \label{fig:dataset}
\end{figure*}

To alleviate this divergence problem, it was suggested to control the parameters to be initialized and remain below the divergence boundary---an analytically derived relationship between $\delta$, $b_c$, and $\omega$ that ensures convergence:
\begin{equation}\label{equation:sust}
    p(\omega) = \frac{-1 + \sqrt{1 - (\delta \, \omega)^2}}{\delta}
\end{equation}
\autoref{figure:simulations} (right) shows the parameter space of the BRF neuron, where any $\omega,~b$ combination below the boundary leads to a stable system. Combined with a trainable b-offset $b' > 0$ to ensure flexibility, where $b_c = p(\omega) - b'$ is constant throughout one sequence length, it leads to the final equation of $b(t)$:
\begin{equation}
b(t) = p(\omega) - b' - q(t)
\end{equation}
The key formulations of the BRF neuron are outlined in \autoref{alg:brf}. The algorithmic representation uses time-discrete tensor operations, denoted with superscripts indicating the time index. Additionally, the transition from time $t$ to $t+1$ is interpreted as a time delay of $\delta$. The BRF-RSNN was optimized for the following sequence learning tasks: Sequential MNIST (S-MNIST), permuted S-MNIST \citep{deng2012mnist}, ECG QT database \citep{qtdb}, and the Spiking Heidelberg dataset (SHD; \citeauthor{cramer2020heidelberg}, \citeyear{cramer2020heidelberg}), shown in \autoref{fig:dataset}. 5 runs of the best single-layered recurrent BRF model are saved with leaky integrator output neurons using BPTT while applying the Adam \citep{kingma2014adam} optimizer \citep{higuchi2024}.

\begin{algorithm}[h]
\small
\setlength{\abovedisplayskip}{0pt}
\setlength{\belowdisplayskip}{0pt}
\caption{BRF Forward Pass}
\label{alg:brf}
$\mathbf{b}^{t} = p(\boldsymbol{\omega}) - \mathbf{b}' - \mathbf{q}^{t-1}$ \Comment{Dampening factor with smooth reset}\\
$\mathbf{u}^t = \mathbf{u}^{t - 1} + \delta ((\mathbf{b}^t + i \boldsymbol{\omega}) \, \mathbf{u}^{t-1} + \mathbf{x}^{t})$ \Comment{Membrane potential update}\\
$\boldsymbol{\vartheta}^{t} = \vartheta_c + \mathbf{q}^{t-1}$ \Comment{Threshold with refractory period}\\
$\textbf{z}^t = \Theta (\Re(\textbf{u}^t) - \boldsymbol{\vartheta}^t)$ \Comment{Spike outcome}\\
$\mathbf{q}^t = \gamma \mathbf{q}^{t-1} + \mathbf{z}^t$ \Comment{Refractory period update}\\
\,\\
$\vartheta_c = 1$, $\gamma=0.9$, and $p(\boldsymbol{\omega}) = \frac{-1 + \sqrt{1 - (\delta \, \boldsymbol{\omega})^2}}{\delta}$\\
\scriptsize($\Re$, $\Theta$, $p$ are applied component-wise.)
\end{algorithm}

The single-layered BRF-RSNN showed higher test accuracy than the best performing deep ALIF-RSNN for three datasets and was on par with one: 99~\% compared to 98.7~\% on S-MNIST, 95~\% compared to 94.3~\% on PS-MNIST and 91.7~\% compared to 90.4~\% on SHD with up to seven times less spikes than ALIF networks \citep{higuchi2024}. 

\begin{figure}[t!]
    \centering
    \includegraphics[width=\linewidth]{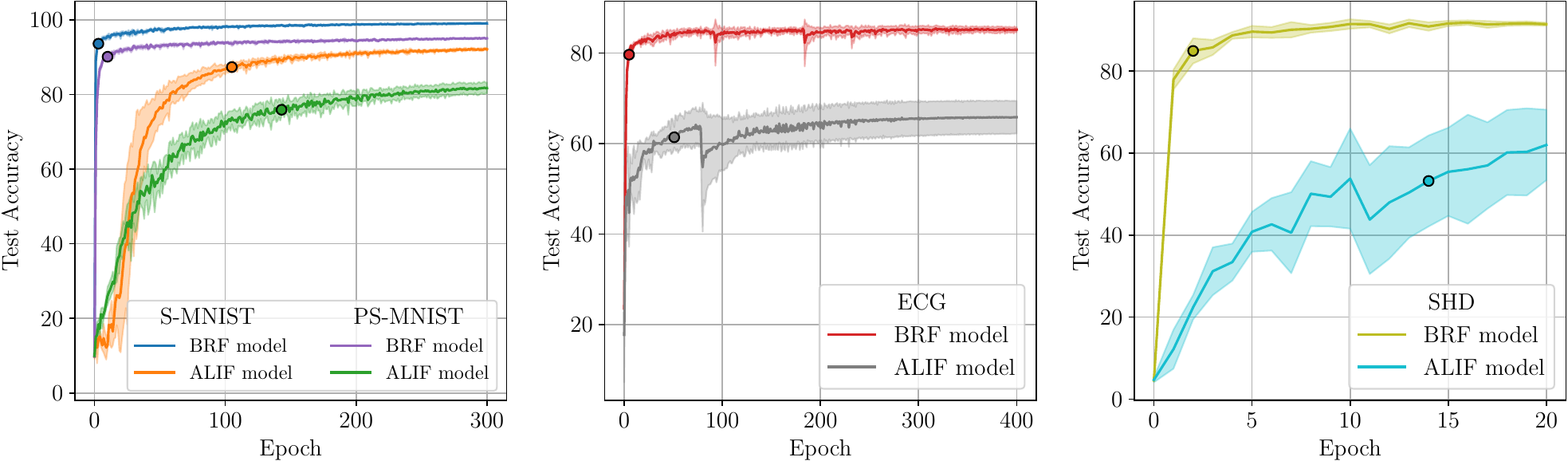}
    \caption{S-MNIST, PS-MNIST, ECG, and SHD learning curve between BRF and ALIF model. Each curve averaged per epoch (solid line) with standard deviation (shaded area) over 5 runs. The dot on the curves depict the point at which 95\,\% of the final accuracy was reached. Adapted from \cite{higuchi2024}.}
    \label{figure:smnist_lc}
\end{figure}

\section{Convergence in BRF-RSNNs}

The convergence results from \cite{higuchi2024}, shown in  \autoref{figure:smnist_lc},  compare the BRF and ALIF model that are both recurrent and single-layered. 
The BRF network consistently reaches 95 \% test accuracy within the first 10 epochs with significant stability compared to the ALIF model. For the BRF networks, the learning rates ranged from 0.075 to 0.1, which is considerably large in the SNN framework, but may have contributed to fast convergence. The BRF neurons did not fire at the beginning of the training phase due to the Xavier-initialized linear weights. Thus, the large learning rate led to fast learning of the weights, enabling the BRF neurons closest to the threshold to spike first. Moreover, the double-Gaussian surrogate gradient function may have contributed as well, such that there was gradient flow even for neurons far from firing \citep{yin2021accurate}. 

\begin{figure}[t!] 
    \centering
    \includegraphics[width=.825\linewidth]{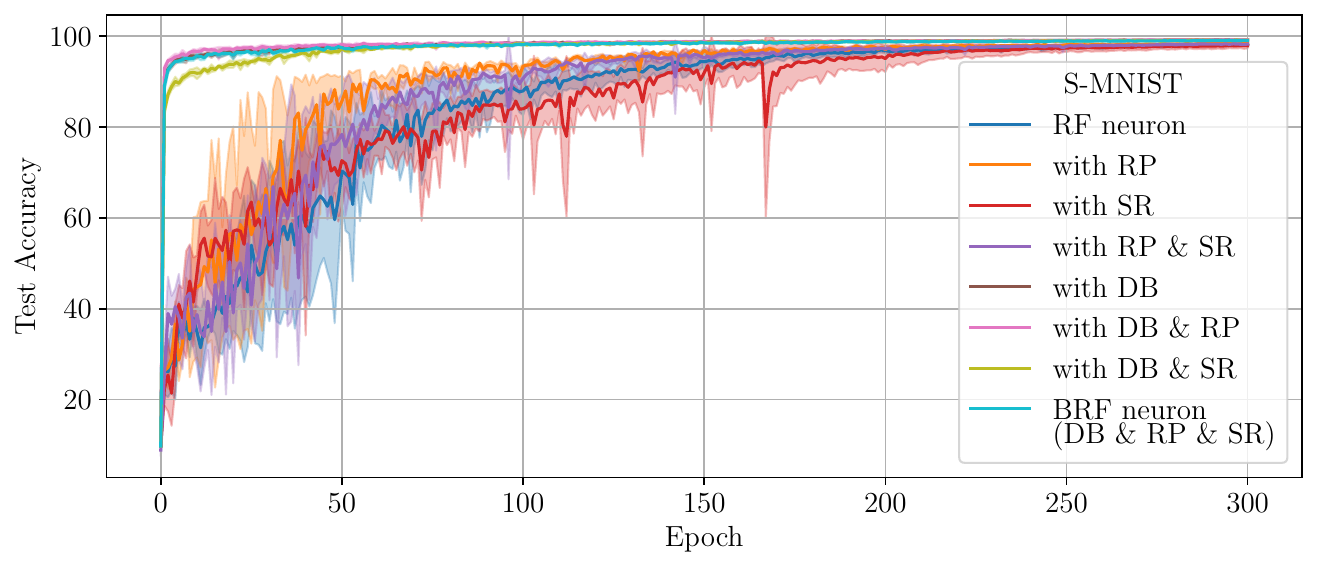}
    \includegraphics[width=.825\linewidth]{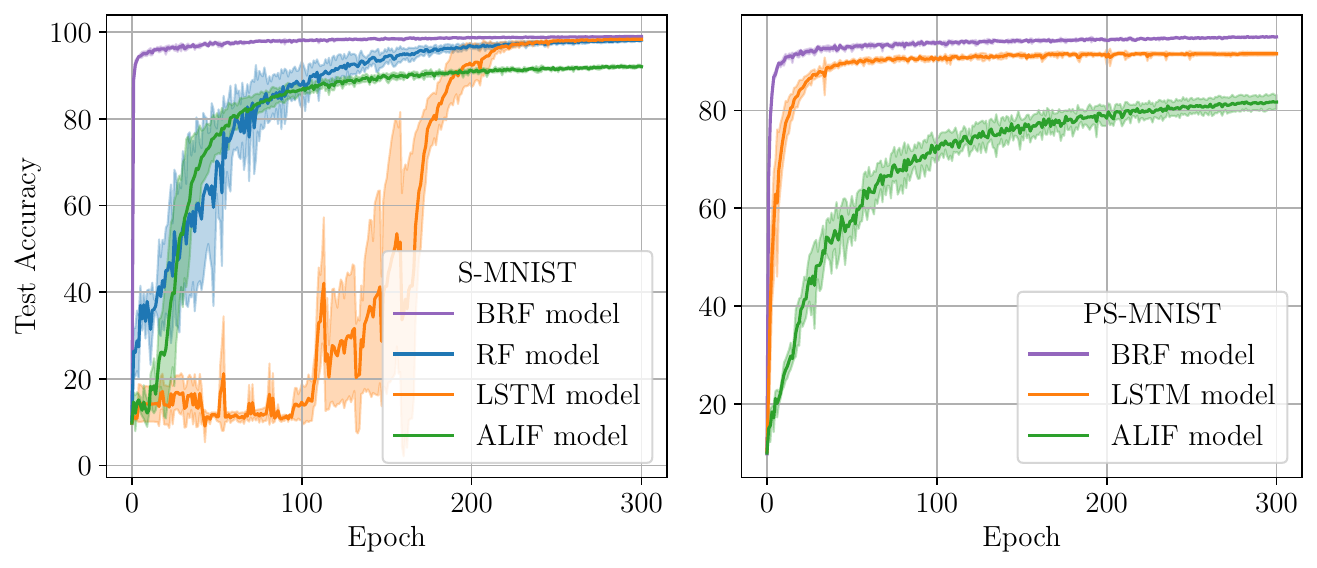}
    \caption{Top: S-MNIST RSNN convergence plot with RF neuon variants. RP: Refractory period; SR: Smooth reset; DB: Divergence boundary. Bottom: S-MNIST (left) and PS-MNIST (right) convergence comparison between (B)RF, standard (non-spiking) LSTM and ALIF model. RF model for PS-MNIST omitted due to loss divergence. Batch size of 256 used for all simulations with 256 hidden units.}
    \label{figure:variant_convergence}
\end{figure}
We examined whether there is a difference in the convergence between basic RF-RSNN and the BRF-RSNN, as well as the combination of the smooth reset, refarctory period and the divergence boundary. All results discussed here are averaged over five runs. The convergence of the variants are compared in \autoref{figure:variant_convergence} (top) for S-MNIST with the same hyperparameters and learning algorithm. Despite the intrinsic limitations mentioned above and the high learning rate of 0.1, the basic RF neuron performed well, presumably due to the stable gradient flow of the near-identity state transition matrix outlined below. More notably, the figure shows a distinctive difference between the implementations with divergence boundary and without, hinting at how the divergence boundary has a prominent impact on the speed and stability of the BRF network. 

Also note that the convergence of a BRF-RSNN is even significantly better than a non-spiking recurrent ANN, namely, an LSTM reference model \citep{hochreiter1997long}, as shown in \autoref{figure:variant_convergence} (bottom).

\subsection{Error landscape}

To further understand the reason for such fast convergence of the BRF neuron, we computed the error landscape of the S-MNIST dataset for the single-layered RF, BRF, and ALIF-RSNNs by means of the following equation:
\begin{equation}
  f(\alpha, \beta) = \mathcal{L}(\theta^* + \alpha \eta + \beta \xi)  
\end{equation}
for $\alpha = \{-1, -0.96 \cdots, 0.96, 1\}$ and $\beta = \{-1, -0.96 \cdots, 0.96, 1\}$ with the filter-wise normalization method for randomly normal distributed direction vectors $\eta$ und $\xi$ \citep{li2018visualizing}.

\begin{figure}[b!]
    \centering
    \includegraphics[width=\linewidth]{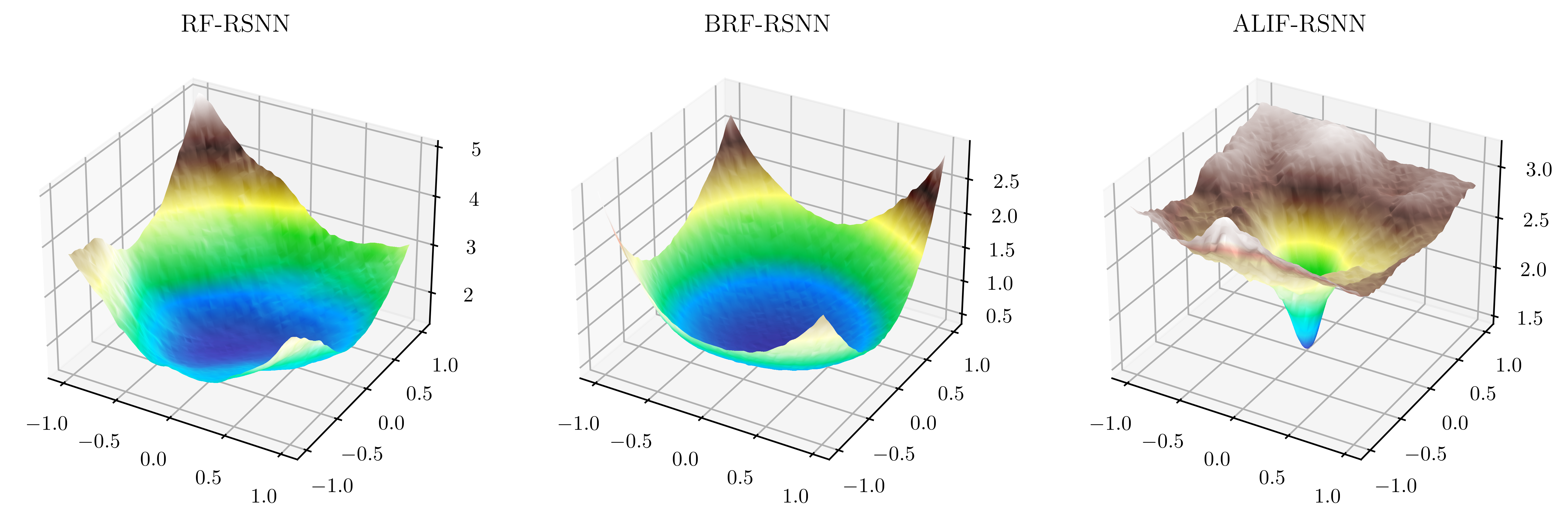}\\
    \includegraphics[width=\linewidth]{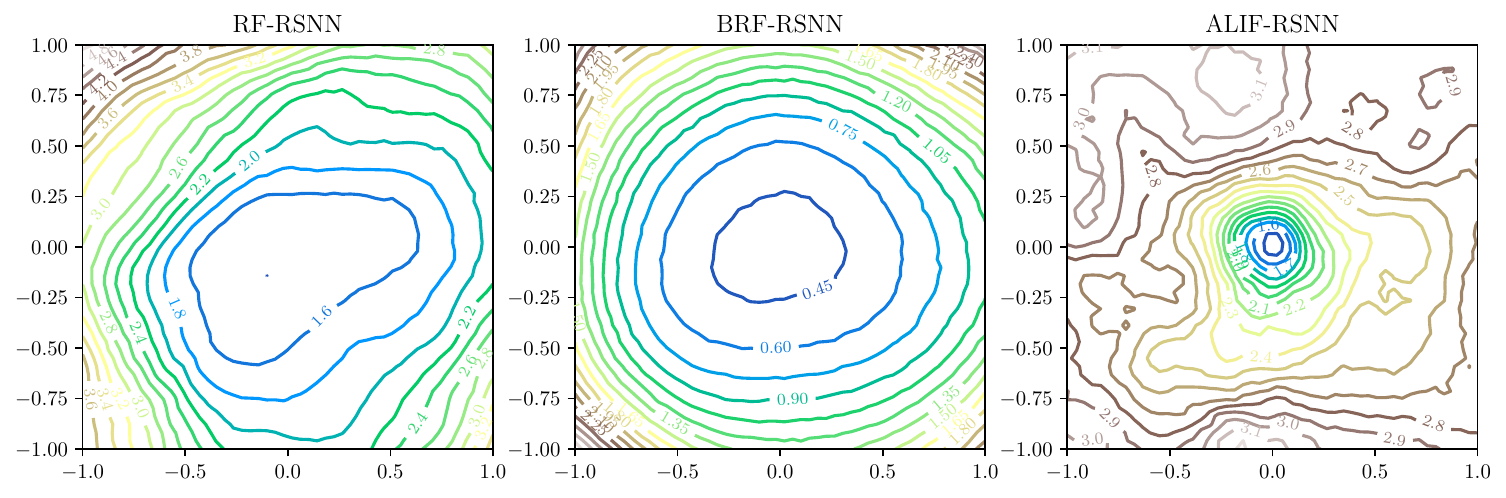}
    \caption{Error landscape plots for RF, BRF, and ALIF network on the S-MNIST dataset. Top: error surface plots. x and y axis correspond to $\alpha$ and $\beta$ and the z axis $f(\alpha, \beta)$. Bottom: the corresponding error contour plots. Note that the value range and hence the coloring does not align across the diagrams for better visualization.}
    \label{figure:error-landscape}
\end{figure}

We applied the equation to the models saved after reaching approximately 50~\% validation accuracy, as the landscape would reflect the nature of the neurons during the major convergence phase. To explore the affect of the BRF and ALIF units on the error, we considered only the input and recurrent connections as $\theta^*$ and left the remaining parameters constant. The respective surface plot as well as the contour plots are shown in \autoref{figure:error-landscape}. 

The BRF network shows smooth near-convex structure without a sign of chaotic behavior compared to the ALIF network with a rough error landscape. Such a flat landscape indicates better generalization and is less prone to over-fitting, as the adjustment of the parameters results in slight changes to the error \citep{li2018visualizing}. The wide basin may also explain the consistent results over multiple runs. On another note, the basic RF network error landscape shows a rougher surface, more narrow basin and higher generalization error than the BRF network, although more smooth and convex than the ALIF counterpart. The convergence of the ALIF network is not stable, as the landscape shows a relatively sharp valley, which highlights the difficulty in optimization, more prone to over-fitting and worse generalization, as also observed in the generally higher test error compared to the BRF network. 

\subsection{Gradient flow analysis}

It seems plausible that the favorable error landscape is a crucial aspect of the superior convergence behavior of BRF neurons over ALIF but also vanilla RF neurons. In the following, we support these empirical findings with a formal investigation of the gradient flow within the BRF model.  

When splitting \autoref{eq:mem_update} into two separate real-valued equations, we obtain:
\begin{align}
u^{t} &= u^{t-1} + \delta(b u^{t-1} - \omega v^{t-1} + x_r^{t}) \\
v^{t} &= v^{t-1} + \delta(\omega u^{t-1} + b v^{t-1})\\
\end{align}
which we can express as a vector:
\begin{align}
\mathbf{s}^{t} &= \begin{bmatrix} u^{t} \\ v^{t} \end{bmatrix}
\end{align}

\begin{figure}[b!]
    \centering
    \includegraphics[width=0.5\linewidth]{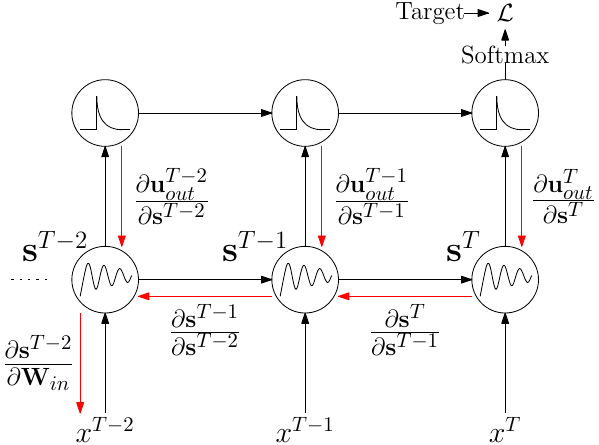}
    \caption{Simplified representation of BPTT for a single BRF neuron with LI output neurons. $\mathbf{s}^t$ denotes the state of the BRF neuron at time $t$. External recurrencies from the outer BRF-RSNN are left out for simplification.}
    \label{figure:grad-flow}
\end{figure}

In the following, we uncover the local state change dependency, that is, we derive the resonator state $\mathbf{s}^{t}$ with respect to the previous state $\mathbf{s}^{t-1}$. Note that $u$ and $v$ additionally influence themselves through the explicit recurrent connections in the network, that is, the output of the resonator is part of the input to the resonator in the next time step. For simplicity, we ignore these outer recurrencies.

We get:
\begin{equation}
\frac{\partial \mathbf{s}^{t}}{\partial \mathbf{s}^{t-1}} = \begin{bmatrix}
\frac{\partial u^{t}}{\partial u^{t-1}} & \frac{\partial u^{t}}{\partial v^{t-1}} \\
\frac{\partial v^{t}}{\partial u^{t-1}} &
\frac{\partial v^{t}}{\partial v^{t-1}}
\end{bmatrix}
\end{equation}
where
\begin{align}
\nonumber \frac{\partial u^{t}}{\partial u^{t-1}} &= \frac{\partial}{\partial u^{t-1}}\left(
u^{t-1} + \delta(b u^{t-1} - \omega
v^{t-1} + x_r^{t})\right) = 1 + \delta b\\
\nonumber \frac{\partial u^{t}}{\partial v^{t-1}} &= \frac{\partial}{\partial v^{t-1}} \left( u^{t-1} + \delta(b u^{t-1} - \omega v^{t-1} + x_r^{t}) \right) = -\delta \omega\\
\nonumber \frac{\partial v^{t}}{\partial u^{t-1}} &= \frac{\partial}{\partial u^{t-1}} \left( v^{t-1} + \delta(\omega u^{t-1} + b v^{t-1}) \right) = \delta \omega\\
\nonumber \frac{\partial v^{t}}{\partial v^{t-1}} &= \frac{\partial}{\partial v^{t-1}} \left( v^{t-1} + \delta(\omega u^{t-1} + b v^{t-1}) \right) = 1 + \delta b
\end{align}

which results in:
\begin{equation}\label{eq:gradflowmatrix}
\frac{\partial \mathbf{s}^{t}}{\partial \mathbf{s}^{t-1}} =
\begin{bmatrix}
1 + \delta b & - \delta \omega \\
\delta \omega & 1 + \delta b
\end{bmatrix}
\end{equation}

Note that this matrix essentially transfers the gradient signal within the BRF neuron back in time and thus primarily determines the gradient flow, as shown in \autoref{eq:gradflowmatrix}.

The first thing we notice is that with a sufficiently small  $\delta$, we approximately end up with:
\begin{equation}
\frac{\partial \mathbf{s}^{t}}{\partial \mathbf{s}^{t-1}} \approx
\begin{bmatrix}
1  & 0 \\
0  & 1
\end{bmatrix}
\end{equation}
In this case, the gradient state transition matrix, which is multiplied with the back flowing gradient at each time step, provides the characteristics of an identity matrix. This essentially means that the gradient remains unchanged within the membrane circuit during BPTT. In practice, however, $\delta$ is usually not too small (more like $0.01$ or $0.1$) and therefore the gradient flow is predominantly affected by $b$ and $\omega$.

In order to get a deeper understanding of the characteristics of the matrix \autoref{eq:gradflowmatrix}, we investigate its  spectral radius, that is, the largest absolute eigenvalue. Recall that the spectral radius of a (square) matrix determines the convergence property of its power series---and thus its scaling characteristics when applied repeatedly--- and converges (to $\mathbf{0}$) when the spectral radius is smaller than one and diverges when it is larger than one. 

The (complex) eigenvalues $\lambda_1$, $\lambda_2$ of the gradient transition matrix from \autoref{eq:gradflowmatrix} are:
\begin{equation}
    \lambda_{1,2} = \delta b \pm i \delta \omega + 1.
\end{equation}
Since $\vert \lambda_{1} \vert = \vert \lambda_{2}|$ holds, we can simply continue with either of them---so we use $\lambda_{1}$. 

Let us now assume a spectral radius of unity---a condition in which the power series neither converges nor diverges:
\begin{equation}
    1 = \vert \delta b + i \delta \omega + 1 \vert
\end{equation}
Solving this equation for $b$ we finally obtain:
\begin{equation}
 b = \frac{-1 + \sqrt{1 - (\delta \, \omega)^2}}{\delta}
\end{equation}

This is exactly the divergence boundary (cf. \autoref{equation:sust}) regulating the convergence/divergence behavior of the resonator circuit. By restricting the parameter space of $b$, it further ensures the spectral radius to remain $\leq 1$, contributing immensely to the stability of the network. This result also explains why the divergence boundary realised in the BRF neuron improves the error landscape to an extent as shown in comparison to the RF neuron: when we choose $b$ and $\omega$ close to the divergence boundary, the magnitude of the gradient is basically preserved.




\section{Conclusion}

This paper investigated how and why the optimization of networks with BRF neurons leads to considerably fast and smooth convergence. The empirical calculation of the error landscape demonstrated that BRF networks exhibit a smooth, flat, and convex structure conductive for generalization and straightforward optimization, contrary to the well-established ALIF networks which we show exhibits a distinctively more chaotic and cluttered error landscape. Moreover, further simulation of varying RF network implementations unraveled the relevance of the divergence boundary for superior stability and convergence speed. The results are analytically grounded in the stable gradient flow within the divergence boundary-aware resonating circuit, preserving the magnitude of the gradient over time.

\begin{ack}
Sebastian Otte was supported by a Feodor Lynen fellowship of the Alexander von Humboldt Foundation.
\end{ack}


\medskip

\bibliographystyle{apalike}

\def\bibfont{\small}
\bibliography{main}


\end{document}